\pdfoutput=1

\documentclass[11pt]{article}

\usepackage[]{emnlp2021}

\usepackage{times}
\usepackage{latexsym}

\usepackage[T1]{fontenc}

\usepackage[utf8]{inputenc}

\usepackage{microtype}
\usepackage{courier}  
\usepackage{amsmath}
\usepackage{bm}
\usepackage{color}
\usepackage{multirow} 
\usepackage{subfig}
\usepackage{url}
\usepackage{amsfonts}
\usepackage{graphicx} 
\usepackage{graphics}
\usepackage{xcolor}
\usepackage[autolanguage]{numprint}
\usepackage{arydshln}
\usepackage{CJKutf8}
\usepackage{microtype}
\usepackage{soul}
\usepackage{paralist}
\usepackage{todonotes}
\usepackage{tabularx}
\title{Document Graph for Neural Machine Translation}

\author{Mingzhou Xu$^{1}$\thanks{~~Work was done when Mingzhou Xu was interning at Noah’s Ark Lab.}, Liangyou Li$^{2}$, Derek F. Wong$^{1}$, \\\bf Qun Liu$^{2}, $\bf Lidia S. Chao$^{1}$, \\
  $^{1}$NLP$^2$CT Lab, University of Macau \\
  $^{2}$Huawei Noah’s Ark Lab\\
  {\tt nlp2ct.mzxu@gmail.com, \{derekfw,lidiasc\}@um.edu.com} \\
  {\tt \{liliangyou,qun.liu\}@huawei.com}}

\begin{document}
\maketitle
\begin{abstract}
 Previous works have shown that contextual information can improve the performance of neural machine translation (NMT). However, most existing document-level NMT methods only consider a few number of previous sentences. How to make use of the whole document as global contexts is still a challenge. To address this issue, we hypothesize that a document can be represented as a graph that connects relevant contexts regardless of their distances. We employ several types of relations, including adjacency, syntactic dependency, lexical consistency, and coreference, to construct the document graph. Then, we incorporate both source and target graphs into the conventional Transformer architecture with graph convolutional networks. Experiments on various NMT benchmarks, including IWSLT English--French, Chinese-English, WMT English--German and Opensubtitle English--Russian, demonstrate that using document graphs can significantly improve the translation quality. Extensive analysis verifies that the document graph is beneficial for capturing discourse phenomena.
\end{abstract}

\section{Introduction}

Although neural machine translation (NMT) has achieved great success on sentence-level translation tasks, many studies pointed out that  translation mistakes become more noticeable at the document-level~\cite{wang-etal-2017-exploiting-cross,tiedemann-scherrer-2017-neural,zhang-etal-2018-improving,miculicich-etal-2018-document,kuang-etal-2018-modeling,voita-etal-2018-context,laubli-etal-2018-machine,tu2018learning, voita2019good,kim2019and,yang2019enhancing}. They proved that these mistakes can be alleviated by feeding the 
contexts into context-agnostic NMT models.

Previous works have explored various methods to integrate context information into NMT models. They usually take a limited number of previous sentences as contexts and learn context-aware representations using hierarchical networks \citep{miculicich-etal-2018-document,wang-etal-2017-exploiting-cross,tan2019hierarchical} or extra context encoders \citep{jean2015using,zhang-etal-2018-improving,yang2019enhancing}. Different from representation-based approaches, ~\citeauthor{tu2018learning}~\shortcite{tu2018learning} and ~\citeauthor{kuang-etal-2018-modeling}~\shortcite{kuang-etal-2018-modeling} propose using a cache to memorize context information, which can be either history hidden states or lexicons. To keep tracking of most recent contexts, the cache is updated when new translations are generated. Therefore, long-distance contexts would likely be erased.

How to use long-distance contexts is drawing attention in recent years. Approaches, like treating the whole document as a long sentence \cite{junczys2019microsoft} and using memory and hierarchical structures \cite{maruf-haffari-2018-document,maruf2019selective,tan2019hierarchical}, are proposed to take global contexts into consideration. However, \citeauthor{kim2019and}~\shortcite{kim2019and} point out that not all the words in a document are beneficial to context integration, suggesting that it is essential for each word to focus on its own relevant context.

\begin{figure}[t] 
  \centering
   \includegraphics[width=0.46\textwidth]{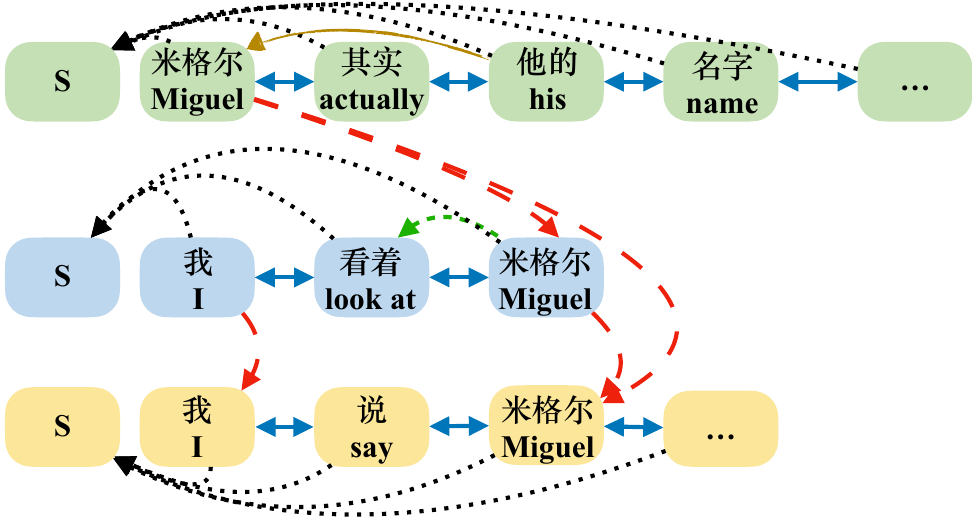}
  \caption{The structure of graph. Solid lines in blue depict adjacency relations. Dash lines in green denote dependency relations. Lexical consistency is represented as dashed lines in red. The brown line means a coreference relation. S denotes Sentence node. We just show aspects of sentences for convenience.\footnotemark}
  \label{fig:exam}
\end{figure}
\footnotetext{Dependency and coreference relations are from Stanford CoreNLP (\url{https://corenlp.run/}).}

To address this problem, we suppose to build a document graph for a document, where each word is connected to those words which have a  direct influence on its translation. Figure \ref{fig:exam} shows an example of a document graph. Explicitly, a document graph 
is defined as a directed graph where: (1) each node represents a word in the document; (2) each edge represents one of the following relations between words: (a) adjacency; (b) syntactic dependency; (c) lexical consistency; or (d) coreference. 

We apply a Graph Convolutional Network (GCN) on the document graph to obtain a document-level contextual representation for each word,  fed to the conventional \textsc{Transformer} model \citep{vaswani2017attention} by additional attention and gating mechanisms.
We evaluate our model on four translation benchmarks, IWSLT English--French (En--Fr) and Chinese--English (Zh--En), Opensubtitle English--Russian (En--Ru), and WMT English--German (En--De). Experimental results demonstrate that our approach is consistently superior to previous works ~\cite{miculicich-etal-2018-document,tu2018learning,zhang-etal-2018-improving,mace2019using,tan2019hierarchical,maruf2019selective} on all the language pairs. 

Contributions of this work are summarized as:
\begin{compactitem}
  \item We represent a document as a graph that connects relevant contexts regardless of their distances. To the best of our knowledge, this is the first work to introduce such graphs into document-level neural machine translation.
  \item We investigate several 
  relations between words to construct document graphs and verify their effectiveness in experiments.
  \item We propose a graph encoder to learn graph representations based on GCN layers with an attention mechanism to combine representations of different sources.
  \item We proposed a context integration method that examined the proposed graph model in different context-aware MT architectures.
\end{compactitem}

\section{Approach}
In this section, we introduce the proposed document graph and model for leveraging contextual information from documents. Firstly, we present a definition of the problem. Then, the construction and representation learning of document graphs are explained in Section~\ref{sec:con-graph} and Section~\ref{sec:doc-encoder}, respectively. Finally, we describe the method of integrating document graphs and model architectures that we use to examine the integration.

\subsection{Problem Definition}

Document-level NMT learns to translate from a document in a source language to a document in a target language. Formally, a source document is a set of $M$ sentences $\mathbf{X}=[X^1,...,X^m,...,X^M]$, where $X^m = [x_1^m,...,x_i^m,...,x^m_{I_m}]$ indicates the $m$th sentence of the document. The corresponding target document is $\mathbf{Y}=[Y^1,...,Y^m,...,Y^M]$, where $Y^m=[y^m_1,...,y^m_j,...,y^m_{J_m}]$ is a translation of the source sentence $X^m$. 

Given the source document to translate, we assume that there is a pair of source and target hidden graphs $G_{\mathbf{X},\mathbf{\hat{Y}}}=\left<G_{\mathbf{X}}, G_\mathbf{\hat{Y}}\right>$ (called document graphs and defined in Section \ref{sec:con-graph}) to help generate the target document. Therefore, the translation probability from $\mathbf{X}$ to $\mathbf{Y}$ can be represented as:
\begin{align}
  &P(\mathbf{Y}|\mathbf{X}) \nonumber\\
        &=\sum_{G_{\mathbf{X},\mathbf{\hat{Y}}}}P(\mathbf{Y}|\mathbf{X},G_{\mathbf{X},\mathbf{\hat{Y}}})P(G_{\mathbf{X},\mathbf{\hat{Y}}}|\mathbf{X})\label{eq:docnmt} \\
        &\propto P(\mathbf{Y}|\mathbf{X},G_{\mathbf{X},\mathbf{\hat{Y}}})\label{eq:docnmt:approx}
\end{align}

Equation~(\ref{eq:docnmt}) is computationally intractable. 
Therefore, instead of considering all possible graph pairs, we only sample one pair of graphs according to the source document resulting in a simplified Equation~(\ref{eq:docnmt:approx}). The construction of source and target graphs are described in Section \ref{sec:con-graph}.



The translation of a document is further decomposed into translations of each sentence with document graphs as context:
\begin{align}
  P(\mathbf{Y}|\mathbf{X}) \approx \prod_{m=1}^{M}P(Y^m|X^m, G_{\mathbf{X},\mathbf{\hat{Y}}}) \label{eq:docnmt:both} 
\end{align}

\subsection{Graph Construction}
\label{sec:con-graph}

Graphs used in this paper are directed, which can be represented as $G=(V, E)$, where $V$ is a set of nodes and $E$ is a set of edges where an edge $e=(u,v)$ with $u,v\in V$ denotes an arrow connection from the node $u$ to the node $v$. 

Our graph contains both word-level and sentence-level nodes. Given a document $\mathbf{X}=[\cdots;x^m_1,\cdots,x^m_{I_m};\cdots]$ where $x^m_i$ is the $i$th ($1\le i\le I_m$) word in the $m$th ($1\le m\le M$) sentence, we construct a document graph with $\sum_{m=1}^{M}I_m$ word-level nodes and $M$ sentence-level nodes. Each word-level node $x_i^m$ in the $m$th sentence is directly connected to the sentence-level node $S_m$. Edges between word-level nodes are determined by intra-sentential and inter-sentential relations. Figure \ref{fig:exam} shows an example document graph. Note that not all edges are depicted for simplicity.

\paragraph{Intra-sentential Relations} provide links between words in a sentence $X^m=x^m_1,\cdots,x^m_{I_m}$. These links are relatively local yet informative and help understand the structure and meaning of the sentence. In this paper, we consider two kinds of intra-sentential relations:
\begin{compactitem}
\item {\bf Adjacency} provides a local lexicalized context that can be obtained without resorting to external resources and has been proven beneficial to sentence modeling \cite{yang2018modeling,xu2019leveraging}. For each word $x^m_i$, we add two edges $(x^m_i, x^m_{i+1}\}$ and $(x^m_i, x^m_{i-1}\}$. This means we add links from the current word to its adjacent words. 
\item {\bf Dependency} directly models syntactic and semantic relations between two words in a sentence. Dependency relations not only provide linguistic meanings but also allow connections between words with a longer distance. Previous practices have shown that dependency relations enhance representation learning of words \cite{marcheggiani2017encoding,strubell2018linguistically,lin2019deep}. Given a dependency tree of the sentence and a word $x^m_i$, we add a graph edge $(x^m_i, x^m_j)$ if $x^m_i$ is a headword of $x^m_j$.
\end{compactitem}

\paragraph{Inter-sentential Relations} allow links from one sentence $X^m=x^m_1,\cdots,x^m_{I_m}$ to another following sentence $X^n=x^n_1,\cdots,x^n_{I_n}$. These relations provide discourse information, which is important for capturing document phenomena in document-level NMT \cite{tiedemann-scherrer-2017-neural,voita-etal-2018-context}. Accordingly, we consider two kinds of relations in our document graph:
\begin{compactitem}
    \item {\bf Lexical consistency} considers repeated and similar words across sentences in the document, which reflects the cohesion of lexical choices. In this paper, we add edges $\{(x^m_i,x^n_j)\}$ if $x^m_i=x^n_j$ or $\mathrm{Lemma}(x^m_i)=\mathrm{Lemma}(x^n_j)$. Namely, the exact same words and words with the same lemma in the two sentences are connected in the graph.
    \item {\bf Coreference} is a common phenomenon in documents and exists when referring back to someone or something previously mentioned. It helps understand the logic and structure of the document and resolve the ambiguities. In this paper we add a graph edge $(x^m_i, x^n_j)$ if $x^m_i$ is a referent of $x^n_j$ given by coreference resolution.
\end{compactitem}
Inter-sentential relations also exist between words in the same sentence, where $m=n$.

\paragraph{Source and Target Graphs} In this paper, we construct a source graph directly from a source document using the method mentioned above. The target graph is built incrementally during inference, i.e., translations of previous sentences in the same document are used as target context. For simplicity, each target context sentence is treated as a fully connected graph and encoded independently by the graph encoder.


\begin{figure}[t] 
  \centering
   \includegraphics[width=0.35\textwidth]{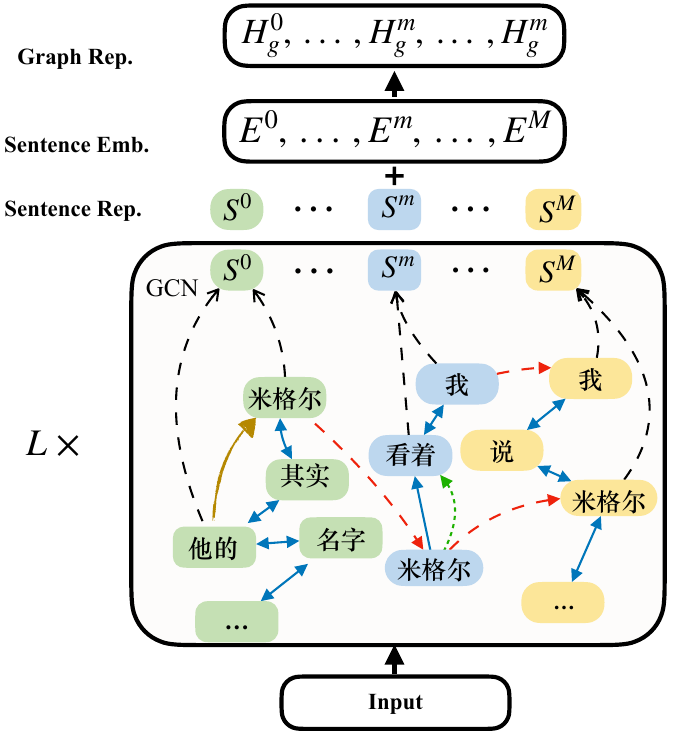}
  \caption{ Illustration of the proposed document graph encoder. $L$ in this paper is set to 2.}
  \label{fig:graph}
\end{figure}

\begin{figure*}[t] 
  \centering
   \includegraphics[width=0.8\textwidth]{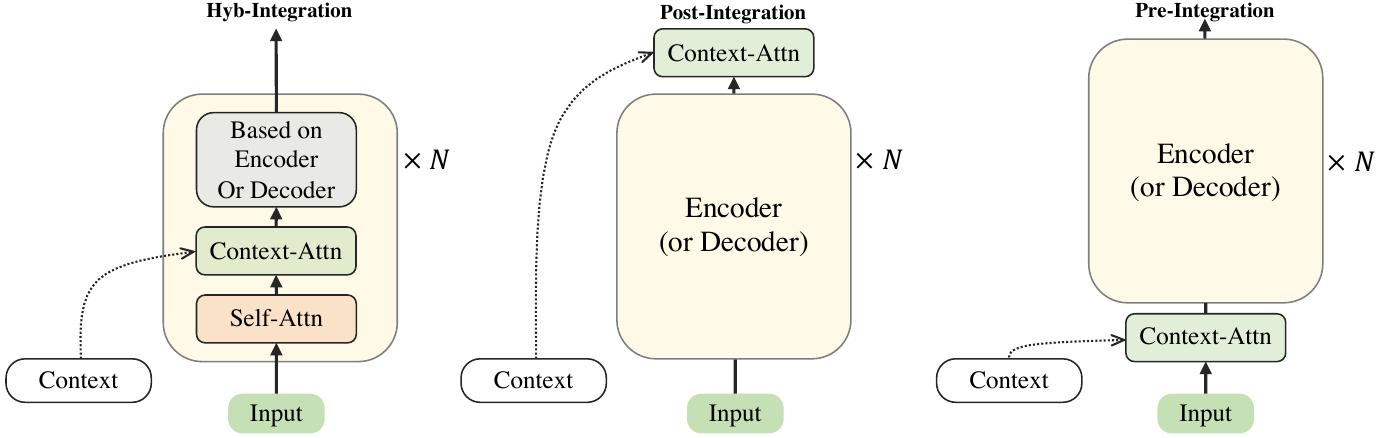}
  \caption{Illustration of the examined architecture. The context information is integrated with a Context-Attn mechanism. Hyb-integration is adding the Context-Attn inside each encoder layer. Post- and Pre-integration is aggregating after and before the encoder, respectively. 
  $N$ in this paper is 6. We only apply source context to the encoder and target context to the decoder, when the contexts are available. Otherwise, we follow the setting of existing works. We share the graph encoder for both source and target graph. Details are shown in Supplementary.}
  \label{fig:arch}
\end{figure*}
\subsection{Document Graph Encoder}
\label{sec:doc-encoder}
As the document is projected into a document graph, a flexible graph encoder is required to encode the complex structure. Previous studies verified that GCNs can be applied to encode linguistic structures such as dependency trees~\cite{marcheggiani2017encoding,bastings2017graph,koncel2019text,huang2020knowledge}. In this paper, we follow previous practices to use stacked GCN layers as the encoder of document graph with considerations on edge directions.

\paragraph{Graph Convolutional Networks}
GCNs are neural networks operating on graphs and aggregating information from immediate neighbors of nodes. Information of longer-distance nodes is covered by stacking GCN layers. Formally, given a graph $G(V,E)$, the GCN network first projects the nodes $V$ into representations $H^0 \in \mathbb{R}^{I \times d}$, where $d$ stands for hidden size and $I=|V|$. Node representations $H^l$ of the $l$th layer can be updated as follows:
\begin{align}
  H^{l+1} = \sigma(D^{-\frac{1}{2}}AD^{-\frac{1}{2}}(W^{l+1}H^l+B^{l+1}))
\label{eq:4}
\end{align}
where $\sigma$ is the sigmoid function and $W^{l+1}\in \mathbf{R^{d \times d}},B^{l+1}\in \mathbf{R^{d}}$ are learnable parameters, $A \in \mathbf{R^{I \times I}}$ is an adjacency matrix that stores edge information: 
\begin{align}
  A(i,j)= 
\begin{cases}
  1 ,& \exists(v_i,u_j) \in E ,\\
  0 ,              & \mathrm{otherwise}.
\end{cases}
\label{eq:5}
\end{align}
The degree matrix $D \in \mathbf{R^{I \times I}}$ is assigned to weight the expected importance of a current node based on the number of input nodes, which can be calculated with the adjacency matrix:
\begin{align}
  D(i,j)= 
\begin{cases}
  \sum_{j'=1}^{I}A(j', i),& i= j ,\\
  0 ,              & \mathrm{otherwise}.
\end{cases}
\label{eq:5.1}
\end{align}

\paragraph{Fusion of Edge Information}
Equation (\ref{eq:5}) only considers input features. 
To fully use direction information in the graph, we apply GCN on different types of edges:
\begin{align}
    \hat{H}_t^{l+1} = \sigma(\hat{D}_t^{-\frac{1}{2}}\hat{A}_t\hat{D}_t^{-\frac{1}{2}}(\hat{W}_t^{l+1}H^l+B_t^{l+1}))
\label{eq:6}
\end{align}
where $t \in \{\mathrm{in},\mathrm{out},\mathrm{self}\}$ represents one of the edge types, i.e., input edges, output edges, or a specific type of self-loop edges. We assume the contributions of the representations learned from a different kind of edges should be different. We then apply a type-attention mechanism, which works better than a linear combination in our experiments,\footnote{We report our experiments in Section 2 of Supplementary.} to combine these representations of different edge types:
\begin{align}
  H^{l+1} &= \sum_{t}\alpha_t \hat{H}_t^{l+1} \\
  \alpha_t &= \mathrm{Softmax}(\frac{H^{l} \hat{H}_t^{l+1}}{\sqrt{d}})
\label{eq:7}
\end{align}
where the $\alpha_t$ are attention weights given by a dot-product attention algorithm
~\cite{vaswani2017attention}. 

\paragraph{Sentence Embedding} After the GCN, we extract the sentence-level nodes $S_m$ as context representation. Since the GCN ignores explicitly positional information between sentences, we add a sentence embedding before integrating the context representation into an encoder or decoder. Figure \ref{fig:graph} shows our graph encoder.

\subsection{Integration of Context Representation}
\label{sec:integration}
Context representation $H_{G}$ from the document graph encoder is treated as a memory and used by an attention mechanism, namely:
\begin{align}
  H_c = \mathrm{Context\text{-}Attn}(X,H_{G},H_{G}) \in \mathbb{R}^{I \times d}
  \label{eq:2.1}
\end{align}

where Context-Attn is a multi-head attention function \cite{vaswani2017attention}. Instead of using the standard residual connection in this sublayer, we adopt a gated mechanism following \citeauthor{zhang2019improving}~\shortcite{zhang2019improving} to dynamically control the influence of context information:
\begin{align}
  \mathrm{Gate}(X, H_c) &= \lambda X + (1-\lambda)H_c \\
  \lambda &= \sigma(W_aX + W_cH_c) 
\label{eq:2.2}
\end{align} 

where $\lambda$ are gating weights, and $\sigma(\cdot)$ denotes the sigmoid function. $W_a$ and $W_c$ are the trainable parameters. In the rest of this paper, we use Context-Attn to denote both the attention and gated residual mechanisms.

In this paper, the Context-Attn sublayer is used in three different ways, as shown in Figure \ref{fig:arch}:
\begin{compactitem}
\item {\bf Hyb-integration}: integrates the contextual information with an additional Context-Attn layer inside each encoder layer~\cite{zhang-etal-2018-improving}. 
  \item {\bf Post-integration}: aggregates the contextual information by adding a Context-Attn layer after the encoder~\cite{tan2019hierarchical,miculicich-etal-2018-document,maruf2019selective}.
  \item {\bf Pre-integration}: interpolates the context representation before the encoder, which can be considered as the hierarchical embedded~\cite{ma-etal-2020-simple}.
\end{compactitem}


\section{Experiments}
 

\paragraph{Data}

We evaluate our approach on translation benchmarks with different corpus size:
(1) IWSLT En--Fr and Zh--En translation tasks~\cite{cettolo2012wit3} with around 200K sentence pairs for training. Following convention~\cite{wang-etal-2017-exploiting-cross,miculicich-etal-2018-document,zhang-etal-2018-improving}, both language pairs take dev2010 as the development set. tst2010 is used for testing on En--Fr and tst2010$\sim$tst2013 on Zh--En. 
(2) Opensubtitle2018 En--Ru translation corpus released by~\citeauthor{voita-etal-2018-context}~\shortcite{voita-etal-2018-context}, which contains 6M sentence pairs for training, among which 1.5M sentence pairs have context sentences.
(3) We adopted the WMT19 document-level corpus published by~\citet{scherrer-etal-2019-analysing} for the En-De translation task. This data contains 2.9M parallel sentences with document boundaries and 10.3M back-translated sentence pairs.

All data are tokenized and segmented into subword units using the byte-pair encoding~\cite{sennrich2015neural}. We apply 32k merge steps for each language on En-Fr, En-Ru, En-De tasks, and 30k for Zh-En task. As a node in a document graph represents a word rather than its subwords, we average embeddings of the subwords as the embedding of the node. The 4-gram BLEU~\cite{papineni2002bleu} is used as the evaluation metric.

\paragraph{Models and Baselines}

Models trained in two stages \cite{jean2015using}:  conventional sentence-level \textsc{Transformer} models (denoted as \textsc{Base}) are first trained with configurations following previous works~\cite{zhang-etal-2018-improving,miculicich-etal-2018-document,voita2019good,vaswani2017attention}. Then, we fix sentence-level model parameters and only train additional parameters introduced by our methods. We set the layers of the document graph encoder to 2 and share their parameters\footnote{Please refer to Supplementary for more details.
}.

\begin{table*}[ht]
  \
  \begin{center}\scalebox{0.9}{
  \begin{tabular}{l|l|l|l|l|l|l|l|l|l|l}
    \multirow {2}{*}{\bf Model}& \multicolumn{2}{c}{\bf En-Fr} & \multicolumn{2}{c}{\bf Zh-En} & \multicolumn{2}{c}{\bf En-DE} & \multicolumn{2}{c|}{\bf En-Ru}& \multirow {2}{*}{\bf Para.~$\bigtriangleup$} &\multirow {2}{*}{\bf Speed} \\
      &$\bigtriangleup$&Test&$\bigtriangleup$&Test&$\bigtriangleup$&Test&$\bigtriangleup$&Test& \\
      \hline\hline 
      \textsc{Base}&$-$& $36.93$& $-$&$17.98$&$-$& $40.67$& $-$&$31.98$ & - & $24.9$k \\
      \hdashline
      \multicolumn{11}{c}{\bf Hyb-Integration}\\
      \hdashline
      \textsc{ Ctx}&$-$&$37.55$&$-$&$18.77$&$-$&$40.95$&$31.27$&$31.95$&$22.06$M&$16.3$k\\
      ~+~\textsc{Src-Graph}&$
      +0.78$&$38.32^\Uparrow$&$+0.89$&$19.66^\Uparrow$&$+0.62$&$41.57^\uparrow$&$+0.92$&$32.87^\Uparrow$&$21.01$M&$17.7$K\\
      ~+~\textsc{Tgt-Graph}&$+1.24$&$\textbf{38.79}^\Uparrow$&$+1.44$&$\textbf{20.21}^\Uparrow$&$+0.89$&$\textbf{41.84}^\Uparrow$&$+0.93$&$\textbf{32.88}^\Uparrow$&$21.01$M&$17.0$K\\
      \hdashline
      \multicolumn{11}{c}{\bf Post-integration}\\
      \hdashline
      \textsc{Hm-gdc}&$-$&$37.42$&$-$&$18.52$&$-$&$40.86$&$-$&	$32.07$&$7.30$~~M&$19.9$k\\
      \textsc{Han}$^\ast$&$-$&$37.70$&$-$&$18.69$&$-$&$41.08$&$-$&$32.36$&$7.36$~~M&$14.4$k\\
      \textsc{Selective$^\ast$}&$-$&$37.95$&$-$&$18.95$&$-$&$41.27$&$-$&$32.54$&$8.39$~~M&$7.7$~~k\\
      ~+~\textsc{Src-Graph}&$+0.03$&$37.98$&$+0.61$&$19.56^\uparrow$&$+0.27$&$41.54$&$-0.07$&$32.47$&$6.27$~~M&$19.7$K\\
      ~+~\textsc{Tgt-Graph}&$+0.40$&$\textbf{38.35}^\uparrow$&$+1.07$&$\textbf{20.02}^\Uparrow$&$+0.62$&$\textbf{41.89}^\uparrow$&$+0.01$&$\textbf{32.55}$&$6.27$~~M&$18.9$K\\
      \hdashline
      \multicolumn{11}{c}{\bf Pre-integration}\\
      \hdashline
      \textsc{Unified}&$-$&$38.02$&$-$&$19.01$&$-$&$41.35$&$-$&$32.44$&$0.01$~~M&$19.6$K\\
      ~+~\textsc{Src-Graph}&$+0.77$&$38.79^\Uparrow$&$+0.99$&$20.00^\Uparrow$&$+0.52$&$41.87$&$+0.45$&$32.89^\uparrow$&$5.27$~~M&$19.7$K\\
      ~+~\textsc{Tgt-Graph}&$+0.97$&$\textbf{38.99}^\Uparrow$&$+1.45$&$\textbf{20.46}^\Uparrow$&$+0.98$&$\textbf{42.33}^\Uparrow$&$+0.47$&$\textbf{32.91}^\uparrow$&$6.27$~~M&$18.5$K\\
  
      \hline
  \end{tabular}}
 
  \caption{Main results (BLEU) on IWSLT Zh--En and EN--FR, WMT19 En--De, and Opensubtitle2018 En--Ru translation tasks. ``$\uparrow/\Uparrow$'' denotes significant improvement~\citep{koehn2004statistical} over the best baseline model with context on each task at $p < 0.05/0.01$, respectively. The models in bold are selected to merge with our document graph methods. ``Para.'' and ``Speed'' indicate the model size (M = million) and training speed (tokens/second), respectively. $\ast$ denotes that the model considers the target context.} 
  \label{tab:main}
  \end{center}
  \end{table*}

To compare our graph-based method with prior works, we reimplement several document-level baselines on the~\textsc{Transformer} architecture and replace their context modules with ours (Please refer to Supplementary on details):
\begin{compactitem}
  \item \textsc{Ctx}~\cite{zhang-etal-2018-improving} employs an additional encoder to learn context representations, which are then integrated by cross-attention mechanisms.
  \item \textsc{Han}~\cite{miculicich-etal-2018-document} uses a hierarchical attention mechanism with two levels (word and sentence) of abstraction to incorporate context information from both source and target documents.
  \item \textsc{Hm-gdc}~\cite{tan2019hierarchical} learns representations with a global context using a hierarchical attention mechanism.
  \item \textsc{Selective}~\cite{maruf2019selective} consider both source and target documents by selecting relevant sentences as contexts from a document.
  \item \textsc{Unified}~\cite{ma-etal-2020-simple} employ the first encoder layer of Transformer to encode the current sentence with context information. Then, the context-aware representation of the current sentence is feed to the transformer model.
\end{compactitem}

\subsection{Overall Results}

Table~\ref{tab:main} shows the overall results on four translation tasks. We find that systems with document graphs achieve the best performance among all context-aware systems on all language pairs with comparable or better training speed. This verifies our hypothesis that document graphs are beneficial for modeling and leveraging the context.
With target graphs, the translation quality in terms of BLEU gets slightly improved, which shows the positive effect of the target context to some extent.
Compared with the corresponding baseline model, our model has a comparable or less number of parameters indicating that the improvements of our method are not because of parameter increments. 

\

\begin{table}[t]
  \begin{center}
  \scalebox{0.9}{
  \begin{tabular}{l|l|l|l}
      \bf Ablation & \bf Model & \bf Dev &\bf Test \\
      \hline\hline 
      &\textsc{Base} & $29.75$ &  $36.93$\\
      \hline
      \multirow{3}{*}{\textbf{Relations}} &    +\textsc{Adjacency}&$30.50$&$37.69$\\ 
      &   +\textsc{Dependency}&$30.75$&$37.81$\\  
      &   +\textsc{Lexical}&$30.68$&$37.78$\\
      &   +\textsc{Coreference}&$30.49$&$37.54$\\
      \hdashline
      \multirow{3}{*}{\textbf{Comp.}} & +\textsc{Intra}&$30.95$&$38.04$\\
      &   +\textsc{Inter}&$30.89$&$37.97$\\
      &  +\textsc{All}&$\mathbf{31.79}$ &$\mathbf{38.94}$  \\
      \hline

  \end{tabular}}
  \caption{Ablation study of source graph variants on IWSLT En-Fr, where \textsc{Lexical} represents ``Lexical consistency''. Comp. represents the complementation.}
  \label{tab:adstudy}
  \end{center}
\end{table}

\begin{table}[t]
  \begin{center}
  \scalebox{0.9}{
  \begin{tabular}{cc|l|l}
       \multicolumn{2}{c|}{\bf Ablation}   & \multirow {2}{*}{\bf BLEU} &\multirow {2}{*}{\bf Speed} \\
      \textbf{word}&\textbf{sentence}&&\\
      \hline\hline 
      $-$&$-$ & $29.75$ & $24.9$K\\
      \hline
      $\surd$&$\times$&$\mathbf{31.79}$ &$16.2$K  \\
      $\times$&$\surd$&$31.66$&$\textbf{17.7}$K \\
      $\surd$&$\surd$&$31.75$&$15.6$K \\
      \hline

  \end{tabular}}
  \caption{Influence of word- and sent-level representations on IWSLT En-Fr.}
  \label{tab:nodes}
  \end{center}
\end{table}

\subsection{Ablation Study}
\paragraph{Edge Relations}
To investigate the influence of the graph construction, we first inspect each kind of edge relation individually by constructing graphs using only one of them. Table~\ref{tab:adstudy} shows that each kind of relation itself improves the translation quality over the \textsc{Base} model, which demonstrates the effectiveness of each selected intra-sentential and inter-sentential relation. 
Combining relations can further improve the system, which achieves the best performance when all relations are considered. These results indicate that the selected relations in this paper are complementary to each other.

\paragraph{Word-level vs. Sentence-level Nodes}
We further examined the influence of the context information at different levels (word- and sentence-level). In this experiment, we tried to use representations of word-level nodes as context. For achieving a better performance, only words in the current sentence are selected. The results are shown in Table \ref{tab:nodes}. We can find that using only representations of sentence-level nodes as context (i.e., default setting) achieves comparable BLEU scores but with a faster training speed.

\paragraph{Sentence Embedding}
Table \ref{tab:sentemb} show the influence of sentence embedding. We can find that using sentence embedding slightly improves the performance (+0.2 BLEU). This is because our graphs are directed where positional information is preserved to some extent.

\begin{table}
  \begin{center}
  \scalebox{0.9}{
  \begin{tabular}{c|l|l}
       \bf Ablation   & \multirow {2}{*}{\bf BLEU} &\multirow {2}{*}{Para.~$\bigtriangleup$} \\
      Sentencen embedding&&\\
      \hline\hline 
      $\surd$&$\mathbf{31.66}$ &$20.21$K  \\
      $\times$ & $31.46$ &$20.00$K \\
      \hline
  \end{tabular}}
  \caption{Influence of sentence embedding on the IWSLT En-Fr benchmark.}
  \label{tab:sentemb}
  \end{center}
\end{table}

\begin{figure*}[ht] 
  \begin{center}
  \subfloat[Context-Distance\label{fig:length}]{\includegraphics[width=0.4\textwidth]{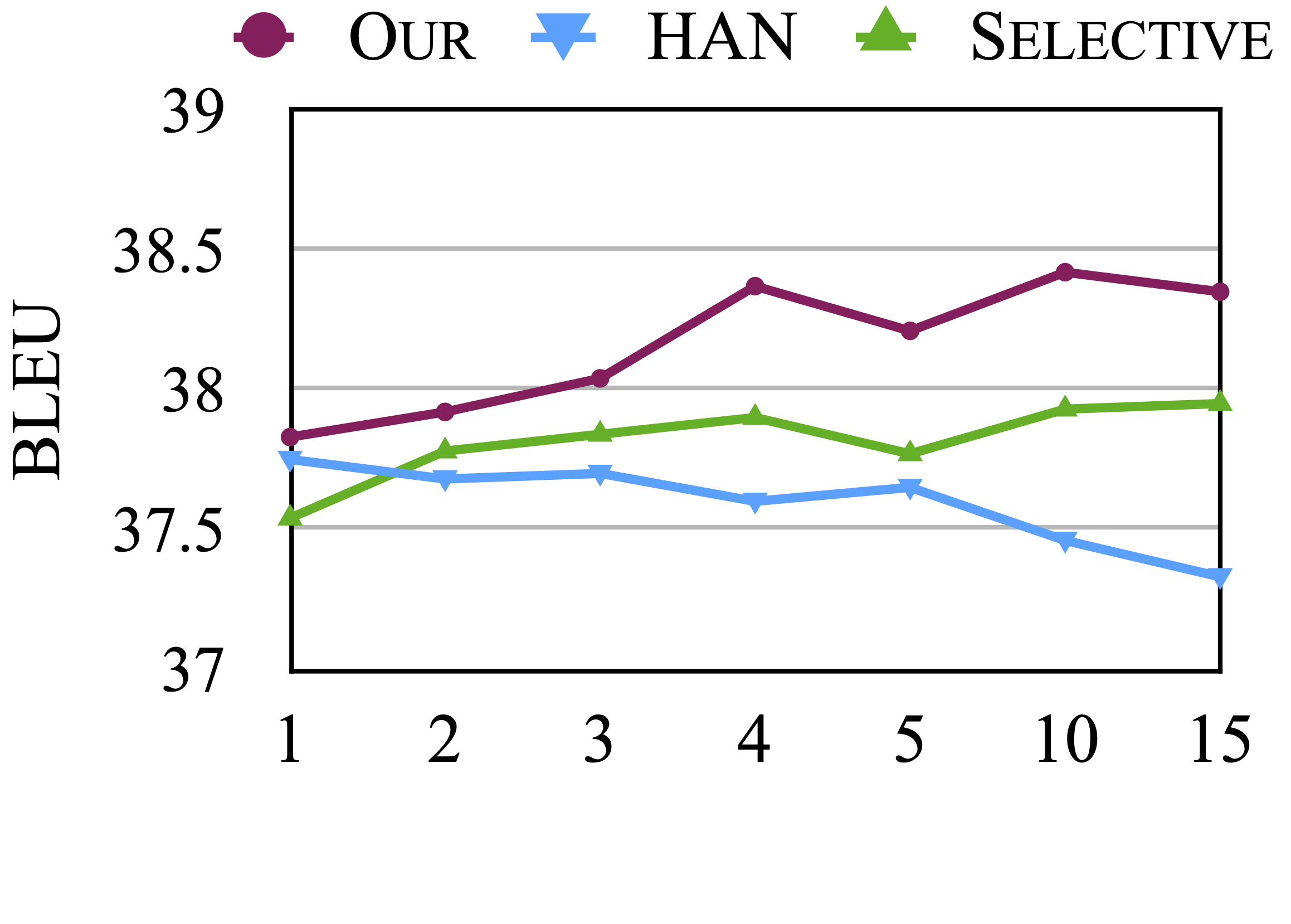}}
  \hspace{0.01\textwidth}
   \subfloat[Document-Size \label{fig:docsize}]{\includegraphics[width=0.42\textwidth]{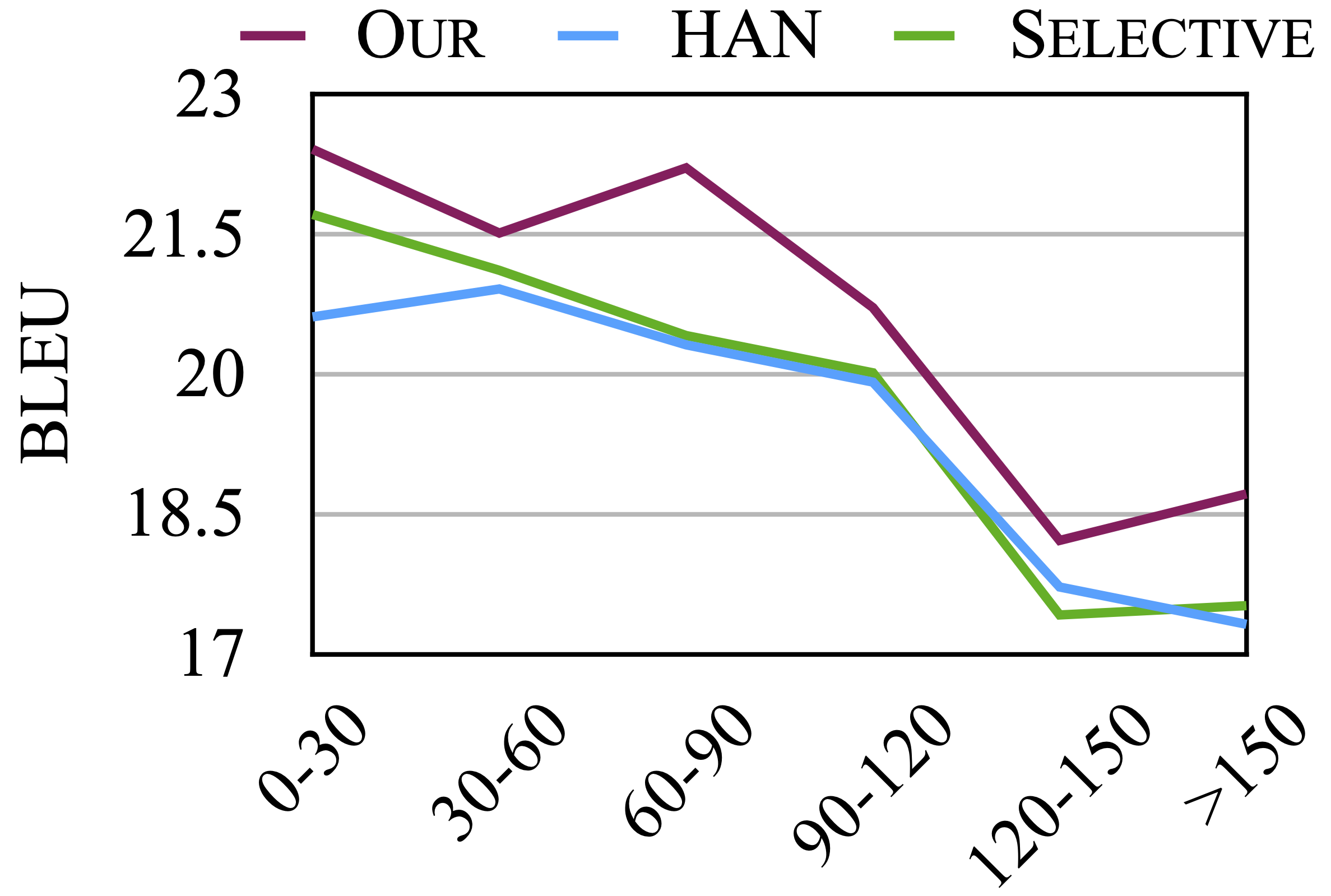}}
  \hspace{0.01\textwidth}
  \end{center}
  \caption{ 
  (a) Visualization of the effectiveness based on the number of sentences considered as contexts in the inference.  The straights are the trend-line of the tested models. (b) Visualization of the effectiveness based on the number of sentences on a document, examined on testing set of Zh-En which contains 56 documents.}
  \label{fig:pos}
  \end{figure*}  

\section{Analysis}
In this section, we analyze the proposed method to reveal its strengths and weaknesses in terms of (1) context distance and its influence; (2) accuracy of dependency tree; (3) changes in document phenomena of translations; and (4) give a case study.

\subsection{Context Distance}

Figure \ref{fig:length} shows the influence of context distance on translation quality. We found that \textsc{HAN} performs worse when increasing the number of context sentences. One possible reason is that sequential structures introduce not only long-distance context but also more irrelevant information. By contrast, our model is getting better while more context is considered. This suggests that graphs help the model focus on relevant contexts regardless of their distance. \textsc{Selective} achieves a lower performance than our model and the gap becomes larger when on longer context, which we surmise is because the attention mechanism has difficulties to differentiate the usefulness of context. This also indicates that the prior knowledge indeed benefits to select relevant context.

Figure~\ref{fig:docsize} shows evaluation results on different document lengths, i.e., the number of sentences in the document. We found that models considering global context (\textsc{Selective} and \textsc{Our}) achieve better results than \textsc{HAN}. \textsc{Our} is consistently better than \textsc{Selective} as well, especially on shorter and longer documents. These results suggest that a global context is beneficial to document-level NMT and appropriate consideration of global context is essential. 

\begin{figure}[ht] 
    \centering
     \includegraphics[width=0.4\textwidth]{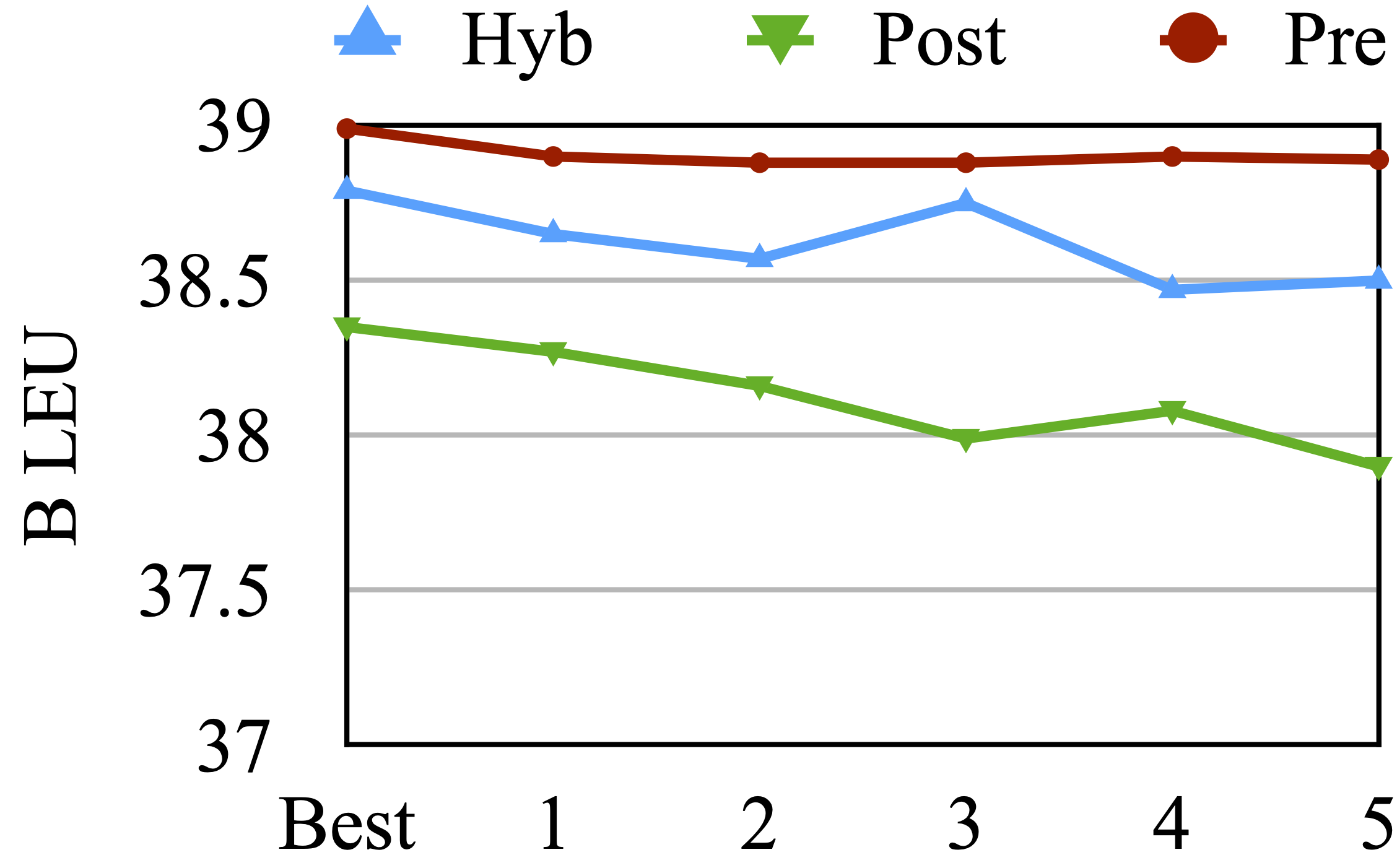}
    \caption{Influence of dependency-tree accuracy on the En--Fr translation task. We examined three different integration methods as described in Section~\ref{sec:integration}. We treat the conversion of k-best results from constituency parser as the dependency tree with decreasing accuracy.}
    \label{fig:dep-acc}
\end{figure}

\subsection{Influence of Dependency-Tree Accuracy}
Figure~\ref{fig:dep-acc} illustrates the influence of accuracy of dependency trees during inference. The {\bf Best} means the best result from the dependency parser. The \textbf{1} to \textbf{5} denote dependency trees converted from the 5-best constituency trees of decreasing accuracy.\footnote{The version of the universal dependency parser in Stanford CoreNLP we used does not support generating n-best results. Therefore, we convert n-best constituent trees into dependency trees.} We find that the performance of our systems with \textbf{Post} and \textbf{Hyb} methods slightly decrease when parsing accuracy becomes lower. However, the \textbf{Pre} method is more robust to parsing accuracy. We attribute this to the fact that integrating the document graph before the encoder leads to more opportunities to resist the noise. 

\begin{table}[t]
  \begin{center}
  \scalebox{0.85}{
  \begin{tabular}{l|ccc|cc}
       \multirow {2}{*}{\bf Model} & \multicolumn{3}{c|}{\bf Consistency}&\multicolumn{2}{c}{\bf Discourse}\\
      & \bf Dex. &\bf Lex. & \bf Ell. & \bf Coref. & \bf Cohe.\\
      \hline\hline 
      \textsc{Base}& $50.0 $& $45.1$& $38.9$& $50.0$& $50.0$ \\
      \textsc{Noise}&$50.0$&$45.2$&$39.6$&$50.5$&$49.5$ \\
      \textsc{HAN}&$60.2$&$57.0$&$64.5$&$55.5$&$53.5$ \\
      \textsc{Selective}&$75.0$&$68.5$&$74.3$&$65.5$&$55.0$ \\
      \textsc{Our}&$\textbf{77.3}$&$\textbf{72.5}$&$\textbf{75.1}$&$\textbf{69.5}$&$\textbf{58.5}$ \\
      \hdashline
      w/o~\textsc{tgt-g}&$60.4$&$63.4$&$59.3$&$57.0$&$55.0$\\
      w/o~\textsc{Intra}&$65.2$&$56.5$&$53.8$&$63.0$&$53.5$\\
      w/o~\textsc{Inter}&$55.4$&$52.7$&$64.1$&$55.0$&$54.5$\\
      \hline
  \end{tabular}}
  \caption{
  Accuracy($\%$) on the Consistency and Discourse test sets. ``\textbf{Dex.}'', ``\textbf{Lex.}'' and ``\textbf{Ell.}'' refer to dexis, lexical cohesion and ellipsis, respectively. \textbf{Coref.} and \textbf{Cohe.} denote the coreference and cohesion tasks, respectively. \textsc{TGT-G} means target graphs. \textsc{INTRA} and \textsc{INTER} are the two group of relations for the graph construction.} 
  
  \label{tab:consis}
  \end{center}
  \end{table}

\subsection{Discourse Phenomena}
We also examine whether our approaches are beneficial to capture discourse phenomena by evaluating our model on the Consistency test set ~\cite{voita2019context} and Discourse test set~\cite{bawden-etal-2018-evaluating}.\footnote{More detailed reports on these tasks are presented in the Supplementary.} 

\paragraph{Test set} The Consistency test set contains three types of tasks on En--Ru: 1) \textbf{Dex.} checks the translation of deictic words or phrases. 2) \textbf{Lex.} focuses on the translation consistency of reiterative phrases. 3) \textbf{Ell.} tests whether models correctly predict ellipsis verb phrases or the morphology of words.

The Discourse test set consists of two probing tasks on En--Fr: 1)~\textbf{Coref.} aims to test whether the gender of an anaphoric pronoun (\textit{it} or \textit{they}) is coherent with the previous sentence. 2)~\textbf{Cohe.} is a set of ambiguous examples whose correct translations rely on the context. 

\paragraph{Result on Discourse Phenomena} As shown in Table~\ref{tab:consis}, all the context-aware models comprehensively improve the performance on discourse phenomena over the context-agnostic~\textsc{Base} model. Results on the the~\textsc{Noise} model~\cite{li2020does} indicate that the improvement is not merely because of robust training.  Compared to prior context-aware models, our model achieves the best accuracy on all tasks. Especially on the Lex., Coref. and Cohe. tasks, our model outperforms others over two points. 
Note that on the ellipsis task graph edges are usually missing for elided verb phrases. For example, given the following source sentence and its context \cite{voita2019good}, the verbs ``told'' and ``did'' are not directly connected in our graph but indirectly connected via the coreference relation of their neighbors ``Nick'' and ``he''. 
Hence, our approach is still slightly better than the best prior method \textsc{Selective}. Directly linking such words may bring further improvements, which we leave for future work. 


\begin{center}
\begin{tabular}{l|l}
\hline
    Context & Nick \textbf{told} you what happened, right? \\
    Source & Yeah, he \textbf{did}. \\
    \hline
\end{tabular}
\end{center}


\paragraph{Analysis on Graphs} We further conduct experiments with the hope of figuring out the influence of graphs on the discourse phenomena, as shown in Table~\ref{tab:consis}. We found that our model with only source graphs (i.e., w/o TGT-G) is consistently better than the \textsc{Base} model on all tasks. Target graphs further improve it to achieve the best performance indicating the importance of target graphs on document-level translation. Both types of relations, \textsc{Inter} and \textsc{Intra}, make significant contributions as well. Their combination brings significant improvement verifying they are complementary to some extent. We also found that compared to \textsc{Intra} relations, \textsc{Inter} relations contribute more on all tasks except the Ell. task. We attribute this to the fact that our document graph contains inter-sentential relations, i.e., lexical consistency and coreference, which directly link relevant contexts for reiterative and deictic words.
\begin{CJK*}{UTF8}{gkai}
\begin{table*}[t]
  \begin{center}
  \scalebox{0.8}{
  \begin{tabularx}{\textwidth}{l|l|X}
    \bf Model & \bf Position & \bf Sentence\\
      \hline\hline 
      \multirow{3}{*}{\textsc{SRC}}&$0$&
      让我们叫他\textcolor{blue}{米格尔}。 其实他的名字就是\textcolor{blue}{米格尔}\\
      &$73$&我一致在脑海中想象类似【帝企鹅日记】的事，我看着\textcolor{blue}{米格尔} \\
      &$74$&我说,"\textcolor{blue}{米格尔},它们飞行150英里来渔场,然后它们晚上再飞150英里回去吗?"\\
      \hline
      \multirow{3}{*}{\textsc{REF}}&$0$& let's call him \textcolor{blue}{miguel}. his name is \textcolor{blue}{miguel}.\\
      &$73$&i was imagining a "march of the penguins" thing,  so i looked at \textcolor{blue}{miguel}.\\
      &$74$&i said, "\textcolor{blue}{miguel}, do they fly 150 miles to the farm, and then do they fly 150 miles back at night?\\
      \hline
      \hline
      \multirow{3}{*}{\textsc{Base}}&$0$& let's call him \textcolor{red}{migoa}. his name is \textcolor{red}{migoingle}.\\
      &$73$&i've always imagined something like a sekhri penguins' diary, and i looked at \textcolor{red}{igel}.\\
      &$74$&i said, "\textcolor{red}{miger}, are they flying 150 miles to fishery, and then they fly 150 miles back at night?"\\
      \hline
      \multirow{3}{*}{\textsc{HAN}}&$0$&let's call him \textcolor{red}{migoa}. his name is \textcolor{red}{migoingle}. \\
      &$73$&i've been thinking about this like 'the penguins diary' in my mind, and i'm looking at \textcolor{blue}{miger}. \\
      &$74$&i said, \textcolor{blue}{miger}, they fly 150 miles to fisheries, and they fly 150 miles at night?\\
      \hline\hline
      \multirow{3}{*}{\textsc{OUR}}&$0$&let's call him \textcolor{blue}{migel}. his name is \textcolor{blue}{migel}.\\
      &$73$&and i've always imagined something like a 'timend penguin diary' in my head, and i'm looking at \textcolor{blue}{migel}. \\
      &$74$& and i said, "\textcolor{blue}{migel}, they fly 150 miles to fisheries, and then they fly 150 miles back at night?\\
  \end{tabularx}}
  \caption{An example of Zh--En task. Compared with \textsc{Base} and \textsc{HAN}, \textsc{Our} system consistently generates ``migel''.}
  \label{tab:case-study}
  \end{center}
  \end{table*}
\subsection{Case-Study}

To verify the long-distance consistency, we perform case studies on the Zh--En task. Table~\ref{tab:case-study} shows an example where a named entity ``米格尔'' (miguel) repeatedly appears in different positions in the document. We first found that both document-level NMT systems, i.e., \textsc{HAN} and \textsc{Our}, generate more consistent translations of the entity than the context-agnostic \textsc{Base} model. Compared with  the \textsc{HAN} model, \textsc{Our} system keeps translating ``米格尔'' into ``migel'', suggesting a more effective capability of handling consistency in long-distance context.

\end{CJK*}


\section{Related work}
In recent years, a variety of studies work on improving document-level machine translation with context. Most of them focus on using a limited number of previous sentences. 
One typical approach is to equip conventional sentence-level NMT with an additional encoder to learn context representations, which are then integrated into encoder and/or decoder ~\citep{jean2015using,zhang-etal-2018-improving,voita-etal-2018-context}. ~\citeauthor{wang-etal-2017-exploiting-cross}~\shortcite{wang-etal-2017-exploiting-cross} and~\citeauthor{miculicich-etal-2018-document}~\shortcite{miculicich-etal-2018-document} adopted hierarchical mechanisms to integrate contexts into NMT models. ~\citeauthor{tu2018learning} ~\shortcite{tu2018learning} and~\citeauthor{kuang-etal-2018-modeling}~\shortcite{kuang-etal-2018-modeling} used cache-based methods to memorize historical translations which are then used in following decoding steps.

Recently, several studies have endeavoured to consider the full document context. ~\citeauthor{mace2019using}~\shortcite{mace2019using} averaged the word embeddings of a document to serve as the global context directly.  ~\citeauthor{maruf-haffari-2018-document}~\shortcite{maruf-haffari-2018-document} applied a memory network to remember hidden states of the document, which are then attended by a decoder.~\citeauthor{maruf2019selective}~\shortcite{maruf2019selective} first selected relevant sentences as contexts and then attended to words in these sentences.~\citeauthor{tan2019hierarchical}~\shortcite{tan2019hierarchical} learned global context-aware representations by firstly using a sentence encoder followed by a document encoder. ~\citeauthor{junczys2019microsoft}~\shortcite{junczys2019microsoft} considered the global context by merely concatenating all the sentences in a document. ~\citet{zheng2020towards} took an additional attention layer to get a representation mixed from the current sentence and whole document. ~\citet{kang2020dynamic} dynamically selected the relevant context from the whole document via a reinforcement learning method. 

Unlike previous approaches, we represent document-level global context in graph encoded by graph encoders and integrated into conventional NMT via attention and gating mechanisms.


\section{Conclusion}
In this paper, we propose a graph-based approach for document-level translation, which leverages both source and target contexts. Graphs are constructed according to inter-sentential and intra-sentential relations. We employ a GCN-based graph encoder to learn the graph representations, which are then fed into the NMT model via attention and gating mechanisms. 
Experiments on four translation tasks and several existing architectures show the proposed approach consistently improves translation quality across different language pairs. Further analyses demonstrate the effectiveness of graphs and the capability of leveraging long-distance context. In the future, we would like to enrich the types of relations to cover more document phenomena.

\section*{Acknowledgements}
This work was supported in part by the Science and Technology Development Fund, Macau SAR (Grant No. 0101/2019/A2), and the Multi-year Research Grant from the University of Macau (Grant No. MYRG2020-00054-FST).

\bibliography{anthology}

\begin{thebibliography}{39}
\expandafter\ifx\csname natexlab\endcsname\relax\def\natexlab#1{#1}\fi

\bibitem[{Bastings et~al.(2017)Bastings, Titov, Aziz, Marcheggiani, and
  Sima'an}]{bastings2017graph}
Joost Bastings, Ivan Titov, Wilker Aziz, Diego Marcheggiani, and Khalil
  Sima'an. 2017.
\newblock \href {https://aclanthology.org/D17-1209} {Graph convolutional
  encoders for syntax-aware neural machine translation}.
\newblock In \emph{ACL}.

\bibitem[{Bawden et~al.(2018)Bawden, Sennrich, Birch, and
  Haddow}]{bawden-etal-2018-evaluating}
Rachel Bawden, Rico Sennrich, Alexandra Birch, and Barry Haddow. 2018.
\newblock \href {https://aclanthology.org/N18-1118/} {{Evaluating Discourse
  Phenomena in Neural Machine Translation}}.
\newblock In \emph{NAACL}.

\bibitem[{Cettolo et~al.(2012)Cettolo, Girardi, and Federico}]{cettolo2012wit3}
Mauro Cettolo, Christian Girardi, and Marcello Federico. 2012.
\newblock \href {https://aclanthology.org/2012.eamt-1.60} {{Wit3: Web inventory
  of transcribed and translated talks}}.
\newblock In \emph{EAMT}.

\bibitem[{Huang et~al.(2020)Huang, Wu, and Wang}]{huang2020knowledge}
Luyang Huang, Lingfei Wu, and Lu~Wang. 2020.
\newblock \href {https://aclanthology.org/2020.acl-main.457/} {Knowledge
  graph-augmented abstractive summarization with semantic-driven cloze reward}.
\newblock In \emph{ACL}.

\bibitem[{Jean et~al.(2015)Jean, Cho, Memisevic, and Bengio}]{jean2015using}
S{\'e}bastien Jean, Kyunghyun Cho, Roland Memisevic, and Yoshua Bengio. 2015.
\newblock \href {https://aclanthology.org/P15-1001/} {{On Using Very Large
  Target Vocabulary for Neural Machine Translation}}.
\newblock In \emph{ACL}.

\bibitem[{Junczys-Dowmunt(2019)}]{junczys2019microsoft}
Marcin Junczys-Dowmunt. 2019.
\newblock \href {https://aclanthology.org/W19-5321/} {{Microsoft translator at
  wmt 2019: Towards large-scale document-level neural machine translation}}.
\newblock In \emph{WMT}.

\bibitem[{Kang et~al.(2020)Kang, Zhao, Zhang, and Zong}]{kang2020dynamic}
Xiaomian Kang, Yang Zhao, Jiajun Zhang, and Chengqing Zong. 2020.
\newblock \href {https://aclanthology.org/2020.emnlp-main.175/} {{Dynamic
  Context Selection for Document-level Neural Machine Translation via
  Reinforcement Learning}}.
\newblock In \emph{EMNLP}.

\bibitem[{Kim et~al.(2019)Kim, Tran, and Ney}]{kim2019and}
Yunsu Kim, Duc~Thanh Tran, and Hermann Ney. 2019.
\newblock \href {https://aclanthology.org/D19-6503/} {{When and Why is
  Document-level Context Useful in Neural Machine Translation?}}
\newblock In \emph{DiscoMT}.

\bibitem[{Koehn(2004)}]{koehn2004statistical}
Philipp Koehn. 2004.
\newblock \href {https://aclanthology.org/W04-3250/} {{Statistical significance
  tests for machine translation evaluation}}.
\newblock In \emph{EMNLP}.

\bibitem[{Koncel-Kedziorski et~al.(2019)Koncel-Kedziorski, Bekal, Luan, Lapata,
  and Hajishirzi}]{koncel2019text}
Rik Koncel-Kedziorski, Dhanush Bekal, Yi~Luan, Mirella Lapata, and Hannaneh
  Hajishirzi. 2019.
\newblock \href {https://aclanthology.org/2020.acl-main.457/} {Text generation
  from knowledge graphs with graph transformers}.
\newblock In \emph{NAACL}.

\bibitem[{Kuang et~al.(2018)Kuang, Xiong, Luo, and
  Zhou}]{kuang-etal-2018-modeling}
Shaohui Kuang, Deyi Xiong, Weihua Luo, and Guodong Zhou. 2018.
\newblock \href {https://aclanthology.org/C18-1050/} {{Modeling Coherence for
  Neural Machine Translation with Dynamic and Topic Caches}}.
\newblock In \emph{Coling}.

\bibitem[{L{\"a}ubli et~al.(2018)L{\"a}ubli, Sennrich, and
  Volk}]{laubli-etal-2018-machine}
Samuel L{\"a}ubli, Rico Sennrich, and Martin Volk. 2018.
\newblock \href {https://aclanthology.org/D18-1512/} {{Has Machine Translation
  Achieved Human Parity? A Case for Document-level Evaluation}}.
\newblock In \emph{EMNLP}.

\bibitem[{Li et~al.(2020)Li, Liu, Wang, Jiang, Xiao, Zhu, Liu, and
  Li}]{li2020does}
Bei Li, Hui Liu, Ziyang Wang, Yufan Jiang, Tong Xiao, Jingbo Zhu, Tongran Liu,
  and Changliang Li. 2020.
\newblock \href {https://aclanthology.org/2020.acl-main.322/} {{Does
  Multi-Encoder Help? A Case Study on Context-Aware Neural Machine
  Translation}}.
\newblock \emph{ACL}.

\bibitem[{Lin et~al.(2019)Lin, Yang, and Lai}]{lin2019deep}
Peiqin Lin, Meng Yang, and Jianhuang Lai. 2019.
\newblock \href {https://www.ijcai.org/proceedings/2019/0707.pdf} {Deep mask
  memory network with semantic dependency and context moment for aspect level
  sentiment classification}.
\newblock In \emph{IJCAI}.

\bibitem[{Ma et~al.(2020)Ma, Zhang, and Zhou}]{ma-etal-2020-simple}
Shuming Ma, Dongdong Zhang, and Ming Zhou. 2020.
\newblock \href {https://aclanthology.org/2020.acl-main.321/} {{A Simple and
  Effective Unified Encoder for Document-Level Machine Translation}}.
\newblock In \emph{ACL}.

\bibitem[{Mac{\'e} and Servan(2019)}]{mace2019using}
Valentin Mac{\'e} and Christophe Servan. 2019.
\newblock \href {https://aclanthology.org/P18-1118/} {{Using whole document
  context in neural machine translation}}.
\newblock In \emph{IWSLT}.

\bibitem[{Marcheggiani and Titov(2017)}]{marcheggiani2017encoding}
Diego Marcheggiani and Ivan Titov. 2017.
\newblock \href {https://aclanthology.org/D17-1159/} {Encoding sentences with
  graph convolutional networks for semantic role labeling}.
\newblock In \emph{EMNLP}.

\bibitem[{Maruf and Haffari(2018)}]{maruf-haffari-2018-document}
Sameen Maruf and Gholamreza Haffari. 2018.
\newblock \href {https://aclanthology.org/P18-1118/} {{Document Context Neural
  Machine Translation with Memory Networks}}.
\newblock In \emph{ACL}.

\bibitem[{Maruf et~al.(2019)Maruf, Martins, and Haffari}]{maruf2019selective}
Sameen Maruf, Andr{\'e}~FT Martins, and Gholamreza Haffari. 2019.
\newblock \href {https://aclanthology.org/N19-1313/} {{Selective Attention for
  Context-aware Neural Machine Translation}}.
\newblock In \emph{NAACL}.

\bibitem[{Miculicich et~al.(2018)Miculicich, Ram, Pappas, and
  Henderson}]{miculicich-etal-2018-document}
Lesly Miculicich, Dhananjay Ram, Nikolaos Pappas, and James Henderson. 2018.
\newblock \href {https://aclanthology.org/D18-1325/} {{Document-Level Neural
  Machine Translation with Hierarchical Attention Networks}}.
\newblock In \emph{EMNLP}.

\bibitem[{Ott et~al.(2019)Ott, Edunov, Baevski, Fan, Gross, Ng, Grangier, and
  Auli}]{ott2019fairseq}
Myle Ott, Sergey Edunov, Alexei Baevski, Angela Fan, Sam Gross, Nathan Ng,
  David Grangier, and Michael Auli. 2019.
\newblock \href {https://aclanthology.org/N19-4009/} {{fairseq: A Fast,
  Extensible Toolkit for Sequence Modeling}}.
\newblock In \emph{NAACL-HLT}.

\bibitem[{Papineni et~al.(2002)Papineni, Roukos, Ward, and
  Zhu}]{papineni2002bleu}
Kishore Papineni, Salim Roukos, Todd Ward, and Wei-Jing Zhu. 2002.
\newblock \href {https://aclanthology.org/P02-1040/} {{BLEU: A Method for
  Automatic Evaluation of Machine Translation}}.
\newblock In \emph{ACL}.

\bibitem[{Scherrer et~al.(2019)Scherrer, Tiedemann, and
  Lo{\'a}iciga}]{scherrer-etal-2019-analysing}
Yves Scherrer, J{\"o}rg Tiedemann, and Sharid Lo{\'a}iciga. 2019.
\newblock \href {https://aclanthology.org/D19-6506/} {{Analysing Concatenation
  Approaches to Document-Level {NMT} in Two Different Domains}}.
\newblock In \emph{DiscoMT 2019}.

\bibitem[{Sennrich et~al.(2016)Sennrich, Haddow, and
  Birch}]{sennrich2015neural}
Rico Sennrich, Barry Haddow, and Alexandra Birch. 2016.
\newblock \href {https://aclanthology.org/P16-1162/} {{Neural Machine
  Translation of Rare Words with Subword Units}}.
\newblock In \emph{ACL}.

\bibitem[{Strubell et~al.(2018)Strubell, Verga, Andor, Weiss, and
  McCallum}]{strubell2018linguistically}
Emma Strubell, Patrick Verga, Daniel Andor, David Weiss, and Andrew McCallum.
  2018.
\newblock \href {https://aclanthology.org/D18-1548/} {{Linguistically-informed
  self-attention for semantic role labeling}}.
\newblock In \emph{EMNLP}.

\bibitem[{Tan et~al.(2019)Tan, Zhang, Xiong, and Zhou}]{tan2019hierarchical}
Xin Tan, Longyin Zhang, Deyi Xiong, and Guodong Zhou. 2019.
\newblock \href {https://aclanthology.org/D19-1168/} {{Hierarchical Modeling of
  Global Context for Document-Level Neural Machine Translation}}.
\newblock In \emph{EMNLP-IJCNLP}.

\bibitem[{Tiedemann and Scherrer(2017)}]{tiedemann-scherrer-2017-neural}
J{\"o}rg Tiedemann and Yves Scherrer. 2017.
\newblock \href {https://aclanthology.org/W17-4811/} {{Neural Machine
  Translation with Extended Context}}.
\newblock In \emph{DiscoMT}.

\bibitem[{Tu et~al.(2018)Tu, Liu, Shi, and Zhang}]{tu2018learning}
Zhaopeng Tu, Yang Liu, Shuming Shi, and Tong Zhang. 2018.
\newblock \href {https://aclanthology.org/Q18-1029/} {{Learning to remember
  translation history with a continuous cache}}.
\newblock In \emph{TACL}.

\bibitem[{Vaswani et~al.(2017)Vaswani, Shazeer, Parmar, Uszkoreit, Jones,
  Gomez, Kaiser, and Polosukhin}]{vaswani2017attention}
Ashish Vaswani, Noam Shazeer, Niki Parmar, Jakob Uszkoreit, Llion Jones,
  Aidan~N Gomez, {\L}ukasz Kaiser, and Illia Polosukhin. 2017.
\newblock \href
  {https://papers.nips.cc/paper/7181-attention-is-all-you-need.pdf} {{Attention
  Is All You Need}}.
\newblock In \emph{NISP}.

\bibitem[{Voita et~al.(2019{\natexlab{a}})Voita, Sennrich, and
  Titov}]{voita2019context}
Elena Voita, Rico Sennrich, and Ivan Titov. 2019{\natexlab{a}}.
\newblock \href {https://doi.org/10.18653/v1/D19-1081} {{Context-Aware
  Monolingual Repair for Neural Machine Translation}}.
\newblock In \emph{EMNLP}.

\bibitem[{Voita et~al.(2019{\natexlab{b}})Voita, Sennrich, and
  Titov}]{voita2019good}
Elena Voita, Rico Sennrich, and Ivan Titov. 2019{\natexlab{b}}.
\newblock \href {https://aclanthology.org/P19-1116/} {{When a Good Translation
  is Wrong in Context: Context-Aware Machine Translation Improves on Deixis,
  Ellipsis, and Lexical Cohesion}}.
\newblock In \emph{ACL}.

\bibitem[{Voita et~al.(2018)Voita, Serdyukov, Sennrich, and
  Titov}]{voita-etal-2018-context}
Elena Voita, Pavel Serdyukov, Rico Sennrich, and Ivan Titov. 2018.
\newblock \href {https://aclanthology.org/P18-1117/} {{Context-Aware Neural
  Machine Translation Learns Anaphora Resolution}}.
\newblock In \emph{ACL}.

\bibitem[{Wang et~al.(2017)Wang, Tu, Way, and
  Liu}]{wang-etal-2017-exploiting-cross}
Longyue Wang, Zhaopeng Tu, Andy Way, and Qun Liu. 2017.
\newblock \href {https://aclanthology.org/D17-1301/} {{Exploiting
  Cross-Sentence Context for Neural Machine Translation}}.
\newblock In \emph{EMNLP}.

\bibitem[{Xu et~al.(2019)Xu, Wong, Yang, Zhang, and Chao}]{xu2019leveraging}
Mingzhou Xu, Derek~F Wong, Baosong Yang, Yue Zhang, and Lidia~S Chao. 2019.
\newblock \href {https://aclanthology.org/P19-1295/} {{Leveraging local and
  global patterns for self-attention networks}}.
\newblock In \emph{ACL2019}.

\bibitem[{Yang et~al.(2018)Yang, Tu, Wong, Meng, Chao, and
  Zhang}]{yang2018modeling}
Baosong Yang, Zhaopeng Tu, Derek~F Wong, Fandong Meng, Lidia~S Chao, and Tong
  Zhang. 2018.
\newblock \href {https://aclanthology.org/D18-1475/} {{Modeling Localness for
  Self-Attention Networks}}.
\newblock In \emph{EMNLP}.

\bibitem[{Yang et~al.(2019)Yang, Zhang, Meng, Gu, Feng, and
  Zhou}]{yang2019enhancing}
Zhengxin Yang, Jinchao Zhang, Fandong Meng, Shuhao Gu, Yang Feng, and Jie Zhou.
  2019.
\newblock \href {https://aclanthology.org/D19-1164/} {{Enhancing Context
  Modeling with a Query-Guided Capsule Network for Document-level
  Translation}}.
\newblock In \emph{EMNLP-IJCNLP}.

\bibitem[{Zhang et~al.(2019)Zhang, Titov, and Sennrich}]{zhang2019improving}
Biao Zhang, Ivan Titov, and Rico Sennrich. 2019.
\newblock \href {https://aclanthology.org/D19-1083/} {{Improving Deep
  Transformer with Depth-Scaled Initialization and Merged Attention}}.
\newblock In \emph{EMNLP-IJCNLP}.

\bibitem[{Zhang et~al.(2018)Zhang, Luan, Sun, Zhai, Xu, Zhang, and
  Liu}]{zhang-etal-2018-improving}
Jiacheng Zhang, Huanbo Luan, Maosong Sun, Feifei Zhai, Jingfang Xu, Min Zhang,
  and Yang Liu. 2018.
\newblock {Improving the Transformer Translation Model with Document-Level
  Context}.
\newblock In \emph{EMNLP}.

\bibitem[{Zheng et~al.(2020)Zheng, Yue, Huang, Chen, and
  Birch}]{zheng2020towards}
Zaixiang Zheng, Xiang Yue, Shujian Huang, Jiajun Chen, and Alexandra Birch.
  2020.
\newblock \href {https://www.ijcai.org/proceedings/2020/0551.pdf} {{Towards
  Making the Most of Context in Neural Machine Translation}}.
\newblock In \emph{IJCAI}.

\end{thebibliography}
\bibliographystyle{acl_natbib}
\newpage
~
\newpage
\appendix

\section{Experiments}
\begin{table*}[ht]
  \begin{center}
  \begin{tabular}{l|l|l|ll|ll|ll}
      \multirow{2}{*}{\bf  Benchmark} &  \multirow{2}{*}{\bf Language} &\textbf{Sent--level} & \multicolumn{2}{c|}{\bf Doc--level} & \multicolumn{2}{c|}{\bf Development} & \multicolumn{2}{c}{\bf testing} \\
      &&&Doc.&Sent.&Doc.&Sent.&Doc.&Sent.\\
      \hline\hline 
      \multirow{2}{*}{IWSLT} \footnote{(\url{https://wit3.fbk.eu/})} &En--Fr&--& $1,823$&$220$K&$8$&$887$&$11$&$1,664$ \\
      &Zh--En&--&$1,718$&$199$K&$8$&$887$&$56$&$5,473$\\
      \hline
      Opensubtitle\footnote{\url{https://github.com/lena-voita/good-translation-wrong-in-context}}&En--Ru&$6.0$M&$1.5$M&$1.5$M&$10$K&$10$K&$10$K&$10$K\\
      \hline
      WMT\footnote{\url{https://github.com/Helsinki-NLP/doclevel-MT-benchmark}}&En--De&$13.2$M&$62,592$&$2.9$M&$236$&$5,168$&$122$&$2,998$\\
      \hline
  \end{tabular}
  \caption{Statistics of the Dataset, where ``Doc.'' is the count of documents and ``Sent.'' denotes the number of sentence pairs.}
  \label{tab:data}
  \end{center}
\end{table*}
\paragraph{Data}

The statistics of the datasets are reported in Table~\ref{tab:data}. For the Chinese language, we segment the data set with the jieba toolkit but the Moses tokenizer.pl for the other languages. WMT19 and Opensubtitle are will pre-processed by ~\citet{scherrer-etal-2019-analysing} and ~\citet{voita-etal-2018-context}.

\begin{table}[ht]
  \begin{center}
  \begin{tabular}{l|r}
      \bf Aggregation  & \bf BLEU \\
      \hline\hline
      \textsc{Gating Units}& $31.41$ \\
      \textsc{Attention}&$\mathbf{31.59}$\\
      \hline
  \end{tabular}
  \caption{Results of aggregation methods in the graph encoder for combining representations learned from different edge directions. \textsc{Gating Units} denotes the weights of summation are calculated by a gating mechanism~\citep{bastings2017graph}. \textsc{Attention} generates weights with an attention mechanism.}
  \label{tab:direct}
  \end{center}
\end{table}

\begin{table}[ht]
    \begin{center}
    \begin{tabular}{c c c}
         \bf \#Layers  & \bf Shared&\bf BLEU \\
        \hline\hline
        1&--&$31.47$\\
        2&--&$31.59$\\
        2&Share&$\bf 31.66$ \\
        3&--&$31.52$\\
        
        \hline
    \end{tabular}
    \caption{Influence of the number of Graph encoder layers used in the graph encoder on IWSLT En--Fr task. }
    \label{tab:layer}
    \end{center}
\end{table}

\begin{table}[ht]
  \begin{center}
  \scalebox{0.9}{
  \begin{tabular}{l|l|l|l}
      \bf Ablation & \bf Model & \bf Dev &\bf Test \\
      \hline\hline 
      &\textsc{Base} & $29.75$ &  $36.93$\\
      &+\textsc{TF-IDF} & $30.63$ &  $37.74$\\
      &+\textsc{All}&$\mathbf{31.66}$ &$38.79$  \\
      \hline

  \end{tabular}}
  \caption{Ablation study of graph variants on the IWSLT En-Fr benchmark,where \textsc{TF-IDF} is the model with the graph constructed by TF-IDF method. \textsc{All} is using the examined relations to construct the graph }
  \label{tab:lower-resouce}
  \end{center}
\end{table}

\begin{table}[ht]
  \begin{center}
  \scalebox{0.9}{
  \begin{tabular}{l|l|l|l}
      \bf Ablation & \bf Model & \bf Dev &\bf Test \\
      \hline\hline 
      &\textsc{Src-graph} & $30.93$ &  $38.32$\\
      &\textsc{Tgt-graph} & $30.79$ &  $38.10$\\
      &\textsc{Both}&$\mathbf{31.66}$ &$38.79$  \\
      \hline

  \end{tabular}}
  \caption{Ablation study of graph variants on the IWSLT En-Fr benchmark,where \textsc{Src-graph} is the model with the source graph. \textsc{Tgt-graph} is only using the target graphas the context.}
  \label{tab:graph-con}
  \end{center}
\end{table}

\paragraph{Settings}

We incorporate the proposed approach into the widely used context-agnostic framework~\textsc{Transformer}~\citep{vaswani2017attention} on \textsc{Fairseq} toolkit~\cite{ott2019fairseq}. The model are trained on V100 GPU.
The conventional context-agnostic~\textsc{Transformer} models are trained with \textsc{BASE} settings. For the IWSLT and Opensubtitle benchmarks, we train the context-agnostic model with 0.2 dropout. The learning rate is set to 0.0007 with 4k warm-up steps.  We set the dropout of the document graph encoder to 0.2, which tuned on validation set. We use approximately 16,000 tokens in a mini-batch for En-Fr, Zh-En, En-Ru, and 32,000 for En--De.  

In decoding, the beam size is set to 4. Following the setting of previous work~\citep{zhang-etal-2018-improving,miculicich-etal-2018-document,voita2019good}, we set the hyper-parameter $\alpha$ of length penalty to 0.6 for En--Fr, En--De, 0.5 for En-Ru and 1 for Zh--En.

\section{Ablation Study}
\paragraph{Graph Encoder}

We extend the GCN-based graph encoder with an attention mechanism to combine different representations, which is different from the gate-based method in previous work \citep{bastings2017graph}. Table \ref{tab:direct} shows that the attention-based aggregation works better in our model. We presume this is because the attention mechanism balances the contributions of different representations. Table~\ref{tab:layer} shows the influence of the graph encoder with various numbers of layers. We found that stacking two graph encoder layers and sharing their parameter obtains the best performance. Further increasing the number of layers does not lead improvement. This finding is consistent with existing works as well~\citep{marcheggiani2017encoding,bastings2017graph}. As shown in Table~\ref{tab:lower-resouce}, we also investigate the traditional TF-IDF construction method, the result indicates that our method is not limited to the examined relations but also works with other graph construction methods.

\paragraph{Graph Contribution} We evaluated the performance of the context form each side. As seen in Table~\ref{tab:graph-con},  only using the source or target side graph shows comparable performance. With both source and target context further improve the translation quality.




\subsection{Discourse Phenomena}

\paragraph{Test set} The consistency test set~\cite{voita2019good} contains four tasks on En--Ru: 1)~\textbf{Deixis} aims to detect the deictic words or phrases whose denotation depends on the context. 2)~\textbf{Lex.C} is a lexical cohesion task, which focuses on the reiteration of named entities. 3)~\textbf{Ell.inf} tests the model on words whose morphological form depends on the context. 4)~\textbf{Ell.VP} is to test whether the model can correctly predict the ellipsis verb phrase in Russian. 
Discourse test set~\cite{bawden-etal-2018-evaluating} consists of two probing tasks on En--Fr: 1)~\textbf{Coref.} aims to test the anaphoric pronoun (\textit{it} or \textit{they}) whose gender is coherent with the previous sentence. 2)~\textbf{Coh.} is a set of ambiguous examples whose correct translations rely on the context. The difference between the \textbf{Cor.} and \textbf{Sem.} is whether the context is correct or not.

Table~\ref{tab:ap-consis} and~\ref{tab:disco} show the details of these two testing sets.

\begin{table}[ht]
  \begin{center}
  \scalebox{0.9}{
  \begin{tabular}{l|c|c|c|c}
      \bf Model & \bf Deixis & \bf Lex.C & \bf Ell.inf & \bf Ell.VP\\
      \hline\hline 
      \textsc{Base}& $50.0 $& $45.1$& $52.8$& $25.0$ \\
      \textsc{Noise}&$50.0$&$45.2$&$53.2$&$26.0$ \\
      \textsc{Ctx}&$57.1$&$48.4$&$73.0$&$58.9$\\
      \textsc{Unified}&$56.7$&$65.2$&$67.9$&$58.3$\\
      \textsc{HAN}&$60.2$&$57.0$&$70.1$&$59.0$ \\
      \textsc{Selective}&$75.0$&$68.5$&$74.0$&$\textbf{74.6}$ \\
      \textsc{Post}&$76.9$&$71.3$&$75.6$&$74.3$ \\
      \textsc{Pre}&$\textbf{77.9}$&$\textbf{74.8}$&$75.9$&$74.1$ \\
      \textsc{Hyb.}&$77.3$&$72.5$&$\textbf{76.3}$&$73.9$ \\
      \hdashline
      w/o~\textsc{tgt-g}&$60.4$&$63.4$&$61.2$&$57.4$\\
      w/o~\textsc{Intra}&$65.2$&$56.5$&$54.5$&$53.1$\\
      w/o~\textsc{Inter}&$55.4$&$52.7$&$65.0$&$63.2$\\
      \hline
  \end{tabular}}
  \caption{
  Accuracy($\%$) on Consistency test sets. \textsc{TGT-G} denotes the target graph. \textsc{INTRA} and \textsc{INTER} is the graph construction method.
}
  \label{tab:ap-consis}
  \end{center}
  \end{table}
  
  \begin{table}[ht]
  \begin{center}
  \scalebox{0.9}{
  \begin{tabular}{l|c|c|c|c}
      \multirow {2}{*}{\bf Model} & \multicolumn{3}{c|}{\bf Coref.($\%$)}&\bf Coh.($\%$)\\
      ~ & \bf ALL & \bf Cor. & \bf Sem. & \bf ALL\\
      \hline\hline 
      \textsc{Base}& $50.0 $& $51.0$& $49.0$& $50.0$ \\
      \textsc{Noise}&$50.5$&$47.0$&$54.0$&$49.5$ \\
      \textsc{Ctx}&$55.0$&$54.5$&$55..5$&$52.0$\\
      \textsc{Unified}&$56.0$&$55.0$&$57.0$&$54.0$\\
      \textsc{HAN}&$55.5$&$57.0$&$54.0$&$53.5$ \\
      \textsc{Selective}&$65.5$&$70.0$&$61.0$&$55.0$ \\
      \textsc{Post.}&$68.0$&$70.0$&$66.0$&$56.5$ \\
      \textsc{Pre}&$\textbf{69.5}$&$\textbf{73.0}$&$66.0$&$\textbf{59.5}$\\
      \textsc{Hyb.}&$\textbf{69.5}$&$70.5$&$\textbf{68.5}$&$58.5$\\
      \hdashline
      w/o~\textsc{tgt-g}&$57.0$&$57.0$&$58.0$&$55.0$\\
      w/o~\textsc{Intra}&$63$&$67.0$&$59.0$&$53.5$\\
      w/o~\textsc{Inter}&$55.0$&$56.0$&$54.0$&$54.5$\\
      \hline
  \end{tabular}}
  \caption{
  Accuracy($\%$) on Discourse test sets. \textsc{TGT-G} denotes the target graph. \textsc{INTRA} and \textsc{INTER} is the graph construction method.}
  \label{tab:disco}
  \end{center}
  \end{table} 
  
\begin{figure*}[ht] 
  \centering
   \includegraphics[width=0.9\textwidth]{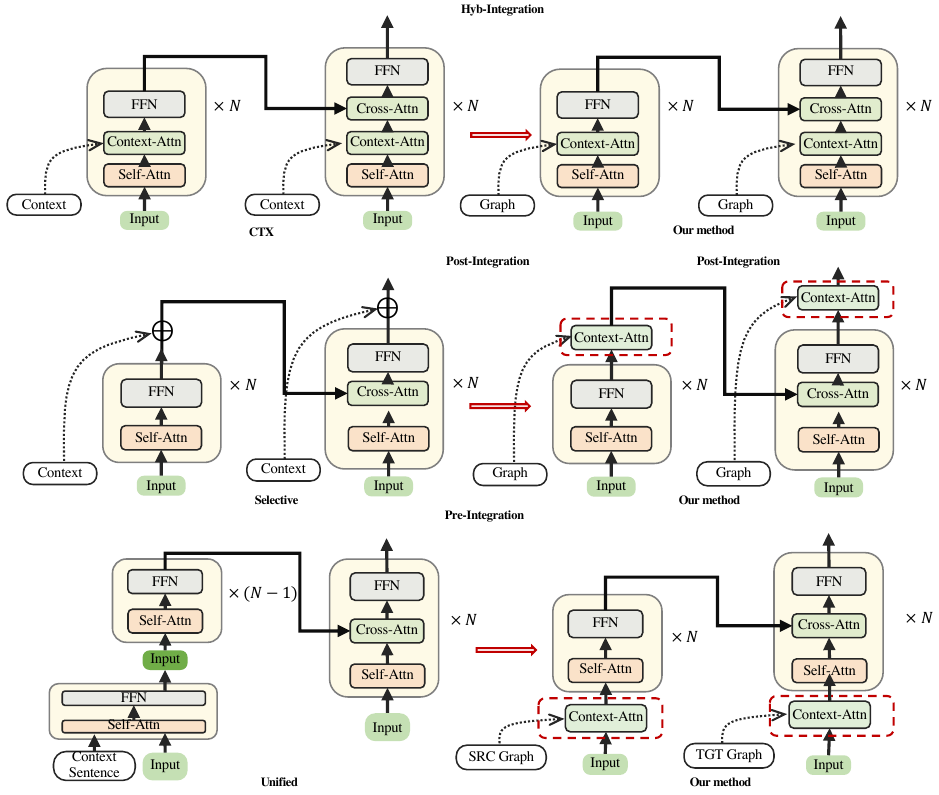}
  \caption{Illustration of the examined architecture. The structures in the red dashed box are the component we added. We didn't modify the basic architecture of the existing works, but take place their context encoder with our graph encoder. Note that the Unified method didn't add the context on the target side.  Therefore, we modified the decoder when we integrate the target graph. }
  \label{fig:ap-arch}
\end{figure*}

\end{document}